\documentclass[lettersize,journal]{IEEEtran}

\usepackage{amsmath,amsfonts}
\usepackage{algorithmic}
\usepackage{array}
\usepackage{textcomp}
\usepackage{stfloats}
\usepackage{url}
\usepackage{verbatim}
\usepackage{graphicx}
\usepackage{algorithm}
\usepackage{cite}
\usepackage{color}
\usepackage{amssymb}
\usepackage{amsmath}
\usepackage{amsthm,amsmath,amssymb}
\usepackage{mathrsfs}
\usepackage{multirow}
\usepackage{booktabs}
\usepackage{balance}
\usepackage[colorlinks=true,linkcolor=black,citecolor=blue,urlcolor=blue,]{hyperref}
\usepackage[doipre={doi:~}]{uri}

\usepackage{subfigure}
\ifCLASSINFOpdf
\else
\fi
\usepackage{amsmath}
\usepackage{multicol}

\usepackage{algorithmic}
\usepackage{algorithm}
\begin{document}
% \title{VoxelMap++: Mergeable Voxel Mapping Method with 3DOF Plane Representation for Online LiDAR(-inertial) Odometry}
\title{VoxelMap++: Mergeable Voxel Mapping Method for Online LiDAR(-inertial) Odometry}

\author{Yifei Yuan, Chang Wu, \IEEEmembership{Member, IEEE}, Yuan You, Xiaotong Kong, Ying Zhang, Qiyan Li
% \thanks{This work was supported by the Science and Technology Project Fund of Sichuan Province, China, under Grant 2018sz0364.}
\thanks{Yifei Yuan, Chang Wu, Xiaotong Kong, Ying Zhang, Qiyan Li are with the Institute of Information and Communication Engineering, University of Electronic Science and Technology of China (UESTC), Chengdu 611731, China (email:yuanyf1998@126.com; changwu@uestc.edu.cn; m13086663208@163.com).}
\thanks{Yifei Yuan and Yuan You have the same contribution to this paper as co-first author. Chang Wu is the Corresponding author.}
\thanks{1 \url{https://www.bilibili.com/video/BV16h4y1r7Cm/} }
\thanks{2 \url{https://github.com/uestc-icsp/VoxelMapPlus_Public} }}
\maketitle

\begin{abstract}
This paper presents VoxelMap++: a voxel mapping method with plane merging which can effectively improve the accuracy and efficiency of LiDAR(-inertial) based simultaneous localization and mapping (SLAM). 
This map is a collection of voxels that contains one plane feature with 3DOF representation and corresponding covariance estimation.
Considering total map will contain a large number of coplanar features (kid planes), 
these kid planes' 3DOF estimation can be regarded as the measurements with covariance of a larger plane (father plane).  
Thus, we design a plane merging module based on union-find which can save resources and further improve the accuracy of plane fitting. 
This module can distinguish the kid planes in different voxels and merge these kid planes to estimate the father plane.
After merging, the father plane 3DOF representation will be more accurate than the kids plane and the uncertainty will decrease significantly which can further improve the performance of LiDAR(-inertial) odometry.
Experiments on challenging environments such as corridors and forests demonstrate the high accuracy and efficiency of our method compared to other state-of-the-art methods (see our attached video$ ^{\hyperlink{target1}{1}}$ ).
By the way, our implementation VoxelMap++ is open-sourced on GitHub$ ^{\hyperlink{target2}{2}}$ which is applicable for both non-repetitive scanning LiDARs and traditional scanning LiDAR.
\end{abstract}


% \fi
\IEEEpeerreviewmaketitle

\section{Introduction}
\label{section1:Introduction}
\IEEEPARstart{R}{ecently}, with the rapid development of 3D LiDAR technology, 
LiDAR(-inertial) odometry has been wildly used in various applications such as 
autonomous vehicles\cite{Baidu1}\cite{Baidu2}, UAVs (Unmanned Aerial Vehicles)\cite{Point-LIO} and
mobile robot\cite{Locus 2.0} because of the robustness and accuracy. 
One important part of LiDAR(-inertial) odometry is the representation of the history map,
almost all state-of-art LiDAR(-inertial) odometry requires the map can provide fast and accurate registration and can increment efficiently during the moving of robots.
The most commonly used mapping method in LiDAR-based simultaneous localization and mapping (SLAM) is KD-Tree point cloud map due to its ease of use, e.g. LOAM \cite{LOAM}, Lego-LOAM\cite{Lego-LOAM}, LIO-SAM\cite{LIO-SAM}, LIO-Mapping\cite{LIO-Mapping}, LINS\cite{LINS}, LILI-OM\cite{LILI-OM}, FAST-LIO\cite{FAST-LIO}.
However, in the process of point cloud registration, it is inevitable to frequently search for K nearest neighbor points to fit the plane which is a time consume task for KD-Tree. 
The time complexity of the nearest find in KD-Tree is O(log(N)) which is harmful to the real-time performance.
Moreover, preserving the uncertainty of point clouds is almost impossible.
Because each point is a random variable of 3DOF, covariance needs to be represented using a 3$\times$3 covariance matrix. 
After covariance estimation, the memory usage of the entire map will expand to four times its original size, which is clearly unaffordable in practice.

\begin{figure*}[t]
\centering
\label{Fig.1}
\includegraphics[width=0.9\textwidth]{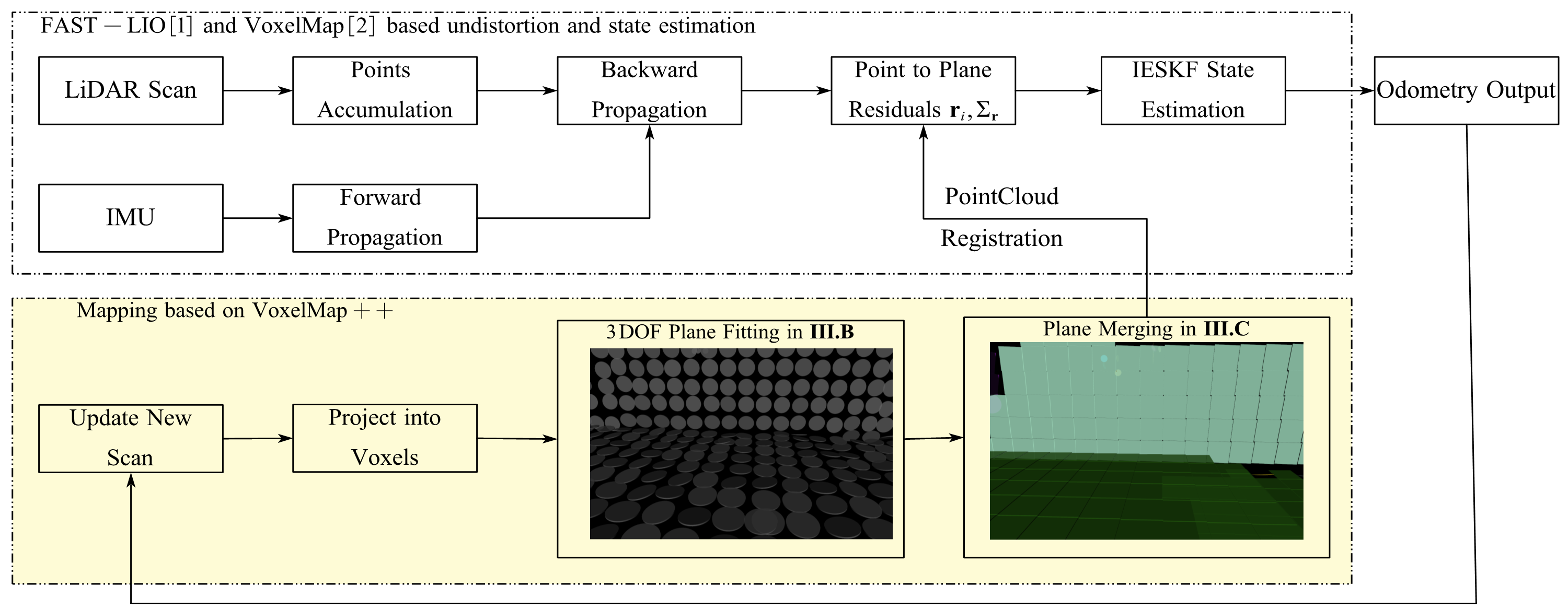}
\caption{
System overview of VoxelMap++, the main contribution of this article is the mapping block denoted in yellow}
\end{figure*}

Fortunately, some voxel mapping methods based on spatial hashing have been proposed by scholars in recent years, such as LIO-Livox, Faster-LIO\cite{Faster-LIO}, BALM\cite{BALM}, BALM 2.0\cite{BALM 2.0}, VoxelMap\cite{VoxelMap}.
The time complexity of registration is O(1), which can guarantee the real-time performance of the system.
In these methods, VoxelMap has excellent accuracy and robustness due to its uncertainty estimation ability about the plane feature.
However, the downside is that it uses 6DOF to represent flat surfaces, which is a waste of memory and increases the time complexity of the total system obviously.
It also does not handle the relationship between different voxels, resulting in high repetition of the same plane in the VoxelMap.
If coplanar voxels can be effectively distinguished, then estimate the large plane formed by these small planes in the voxels. The uncertainty of the entire map will significantly decrease, which will further improve the positioning accuracy of online LiDAR(-inertial) odometry.

To address the above challenges, this paper proposes a novel online mergeable voxel mapping method with 3DOF plane representation termed VoxelMap++.
Concretely, the contributions of this paper include:

\begin{itemize}
   \item  We promote the plane fitting and covariance estimation method in Voxelmap from 6DOF to 3DOF by using least squares estimation. This improvement is from the perspective of engineering implementation, further improving the efficiency of covariance estimation and reducing memory usage, which allows VoxelMap++ to easily adapt to various resource-constrained embedded platforms.
    \item  We propose a novel online voxel merging method by using union-find. Each coplanar feature (kid plane) in the voxel will be regarded as the measurements of a large plane (father plane). The merging module not only arising the plane fitting accuracy and decrease the uncertainty of the total map, but also reduces the memory usage of the map.
    \item  We compared VoxelMap++ with other state-of-art algorithms in various scenarios (structured, unstructured and degenerated) which demonstrates the superiority of the algorithm in accuracy and efficiency.
    \item We make the VoxelMap++ adapt different kinds of LiDARs (multi-spinning LiDARs and non-conventional solid-state LiDARs) and open-sourced with readability and modularity on our GitHub for sharing our findings and making contributions to the community.
\end{itemize}
\section{Replated Work}
\label{section2:Replated Work}
LOAM\cite{LOAM} and its variant\cite{Lego-LOAM} construct the map online by incrementally registering the new scan into point clouds map based on KD-Tree, then perform voxel filtering for downsampling to reduce the number of points in the map.
However, the efficiency of point cloud maps in scan registration is still low because each point-plane matching requires retrieving at least three nearest points to fit the corresponding plane.
As the map increases, the cost of reconstructing KD-Tree also gradually increases. Making KD-Tree based point cloud maps unsuitable for some large-scale scenarios.

To overcome the drawback of map based on point clouds, FAST-LIO2\cite{FAST-LIO2} develops an incremental kd-tree structure\cite{ikd-tree} to organize the point cloud map efficiently; MINS\cite{MINS} maintains a plane patch point cloud to decrease the numbers of point in map, which contain the center point $p$ and Hesse normal $n$ of plane. With the efforts of these researchers, point cloud maps have gradually become practical, but this is accompanied by a series of complex map management methods.

Normal distribution transformation (NDT)\cite{ndt} and its variant\cite{MINS} is the most classic voxel mapping method which divided the positioning spaces into voxels. Each voxel contains a 3D Gaussian distribution by fitting the history point cloud. By calculating the points probabilistic in the voxel, state estimation is formulated as the maximum likelihood. 

Recently, some state-of-art methods (e.g. puma-lio\cite{puma-lio}, Voxelmap\cite{VoxelMap} and BALM\cite{BALM}) add plane assumption in the voxel, and further estimate the covariance of the plane which achieve excellent performance in robustness and efficiency. 
However, these methods ignore the relationship between the neighbor voxels, which leads the entire plane (such as ground) in the map will be divided into many subplanes.
Moreover, due to the limited volume of each voxel, the point clouds used to fitting the plane in the voxel are limited which makes the estimation of plane representation and covariance not accurate enough.

\section{Methodology}
\label{section3:Methodology}

\subsection{System Overview}
\label{section3.1:System Overview}
% This paper will use the notations shown in Table \ref{tab661}.
The pipeline of VoxelMap++ is shown in Fig. \ref{Fig.1}. 
The LiDAR raw points preprocessing method and state estimation method based on iterated error-state Kalman filter are similar to FAST-LIO\cite{FAST-LIO}. 
By the way, the mapping method of this paper can be adapted to other state-of-art LiDAR(-inertial) algorithms no matter it is based on Kalman filter or optimization.

After state estimation, each point in the new scan will be projected into the corresponding voxel, then construct or update the voxel map which is organized by a Hash table (key is voxel id and value is plane fitting results $\mathcal{P}$).
The voxel map construction and update is introduced in \ref{section3.3.2:Voxel Map Construction and Update}. 
These new points will be incrementally used for 3DOF plane fitting and covariance estimation which will be introduced in \ref{section3.2:3DOF Plane Fitting and Covariance Estimation}. 
The complexity of this module will not increase with the number of points in the voxel, because all value used for fitting the plane is the form of summation which can be cached and calculated incrementally.

Then, the converging plane will be used for plane merging which will be introduced in \ref{Plane Merging Based on Union-Find}. 
In this module, kid plane $\mathcal{P}^k$ in the voxels will be merged into father plane $\mathcal{P}^f$ based on union-find. 
Meanwhile, the plane estimation results of $\mathcal{P}^f$ will be more accurate compared with $\mathcal{P}^k$ which will obviously improve the positioning results of LiDAR(-inertial) odometry.

\subsection{3DOF Plane Fitting and Covariance Estimation}
\label{section3.2:3DOF Plane Fitting and Covariance Estimation}
Like other implementations of LiDAR(-inertial) odometry\cite{FAST-LIO}\cite{VoxelMap}\cite{Faster-LIO}\cite{MINS}, we only use the plane features in the environment due to its vast availability. In each voxel, we maintain a 3DOF plane fitting results $\textbf{n} = [a, b, d]^T$ and its covariance $\Sigma_{\textbf{n}}$.
In this subsection, we will illuminate how to fit the plane and estimate its covariance in 3DOF rather than other redundant representations.

Firstly, according to the analysis of the measurement noises for LiDAR sensors in
\cite{Pixel-level}, we can calculate the covariance of each LiDAR points $^L\textbf{p}_i$ in the local LiDAR frame \eqref{eq1}, then using pose estimation results $(^W_L\textbf{R}, ^W_L\textbf{t})$ to transform it to world frame \eqref{eq2}. Thus, we can further estimate the LiDAR points covariance $\Sigma_{^W\textbf{p}_i}$ in world frame \eqref{eq3}. Detailed derivation of \eqref{eq1}\eqref{eq2}\eqref{eq3} can be found in \cite{VoxelMap}.
\begin{gather}
\begin{aligned}
\textbf{A}_i = \left[\begin{matrix}
\boldsymbol{\omega}_i & -d_i \lfloor\boldsymbol{\omega}_i\rfloor_{\times}\textbf{N}(\boldsymbol{\omega}_i)
\end{matrix}\right] \\
\Sigma_{^L\textbf{p}_i} = \textbf{A}_i \left[\begin{matrix}
\Sigma_{d_i} & \textbf{0}_{1\times2} \\
\textbf{0}_{2\times1} & \Sigma_{ \boldsymbol{\omega}_i}
\end{matrix}\right] \textbf{A}_i^T
\end{aligned}
\tag{1}
\label{eq1}
\\
\label{eq2}
\tag{2}
{}^W\textbf{p}_i ={}_{L}^{W}\textbf{R}{}^L\textbf{p}_i + {}^W_L\textbf{t}\\
\label{eq3}
\tag{3}
\Sigma_{^W\textbf{p}_i} = {}_{L}^{W}\textbf{R}(\Sigma_{^L\textbf{p}_i} + \lfloor{}^L\textbf{p}_i\rfloor \Sigma_{{}_{L}^{W}\textbf{R}} \lfloor{}^L\textbf{p}_i\rfloor_{\times}^T){}_{L}^{W}\textbf{R}^T + \Sigma_{{}_{L}^{W}\textbf{t}}
\end{gather}

Assuming a group of coplanar LiDAR points $^W\textbf{p}_i, (i = 1, \cdots, N)$ with covariance $\Sigma_{^W\textbf{p}_i}$ have be determined using \eqref{eq3}.
Then, we leverage the linear least-squares method to calculate the 3DOF representation of plane \cite{MINS}. 
Considering a plane can be extracted from this point cloud parameterized as \eqref{eq4} by normalizing along z-component (main-axis). 
This expression can be singular when the z-component of the plane normal is close to zero.
While this issue can be easily resolved by calculating the projection of the point cloud to each coordinate axis, and the one with the smallest projection variance is the main-axis.
% This expression can be singular when the z-component of the plane normal is close to zero, while this issue can be easily resolved by performing eigenvalue decomposition of pointcloud.
\begin{gather}
\label{eq4}
\tag{4}
ax+by+z+d=0 %\\
% \label{eq5}
% \tag{5}
% axis = \left\{ 
%     \begin{array}{ll}
%         \Sigma y_{i}y_{i}*\Sigma z_{i}z_{i}-\Sigma y_{i}z_{i}*\Sigma y_{i}z_{i}  \\
%         \Sigma x_{i}x_{i}*\Sigma z_{i}z_{i}-\Sigma x_{i}z_{i}*\Sigma x_{i}z_{i}  \\
%         \Sigma x_{i}x_{i}*\Sigma y_{i}y_{i}-\Sigma x_{i}y_{i}*\Sigma x_{i}y_{i} 
%     \end{array}
%         \right\}
\end{gather}

Since all $^W\textbf{p}_i$ are coplanar satisfied \eqref{eq4}, so the least squares optimization function can be constructed as \eqref{eq6}. After a series of identical deformations, a closed-form solution of $\textbf{n}$ can be obtained \eqref{eq7}.
% In VoxelMap, the process to init a plane from points includes doing a singular value decomposition on a covariance matrix to find the eigenvector of the smallest eigenvalue. However, this makes things more complicated than they need to be, for example, a plane can be represented by a normal vector $n=[a, b, c]^{T}$ and a distance \textit{d}:
% While this kind of representation is still redundant because it can be simplified as 3 variables \textit{(a, b, d)} after dividing by \textit{c} in the equation:
% \begin{equation}
%     ax+by+z+d=0
% \end{equation}
% However, this kind of representation is build on the assumption that the simplified \textit{c} is not zero. When the \textit{c} is zero or closed to zero, this simplification will get bad conditioning and thus bad results. To solve this problem, we first choose the main direction of the normal of the plane because there is at least one of the components of the normal must be non-zero if the points do span a plane. So we can decide the representation of the plane by choosing the maximum of \textit{(a, b, c)}, then we discuss the plane with three separate assumptions as below:
% \begin{gather*}
%     ax+by+z+d=0 \\
%     ax+y+bz+d=0 \\
%     x+ay+bz+d=0 
% \end{gather*} 
% For the sake of brevity, only one of the cases will be discussed when seeking \textit{(a, b, d)} and its covariance matrix. Let $
% \textbf{p}_{i}=(x_i, y_i, z_i)$ be the LiDAR point on the plane, N be the number of laser points on the plane, then we construct the least squares problem to solve \textit{(a, b, d)}:
\begin{gather}
\label{eq5}
\tag{5}
\left[\begin{matrix}
x_1 & y_1 & 1 \\
x_2 & y_2 & 1 \\
\vdots & \vdots & \vdots \\
x_n & y_n & 1
\end{matrix}\right]
\textbf{n} = 
\left[\begin{matrix}
-z_1  \\
-z_2 \\
\vdots \\
-z_n
\end{matrix}\right]\\
\label{eq6}
\tag{6}
\textbf{n}= \frac{\mathbf{A}^{*}}{\left|\mathbf{A}\right|}\textbf{e}
\end{gather}

where $\mathbf{A}^{*}$ is the adjugate matrix of $\textbf{A}$, the expression of $\textbf{A}$ and $\textbf{e}$ is shown as 
\begin{equation}
\label{eq7}
\tag{7}
\begin{aligned}
    &\mathbf{A} = \left[\begin{matrix} \sum x_{i}x_{i} & \sum x_{i}y_{i} & \sum x_{i} \\\ \sum x_{i}y_{i} & \sum y_{i}y_{i} & \sum y_{i} \\\ \sum x_{i} & \sum y_{i} & N \end{matrix}\right]
    \\\
    &\textbf{e} = \left[\begin{matrix}-\sum x_{i}z_{i} & -\sum y_{i}z_{i} & -\sum z_{i}\end{matrix}\right]^{T}
\end{aligned}
\end{equation}

Based on the closed-form solution of $\textbf{n}$, 
the uncertainty of the plane is hence 
\begin{equation}
\label{eq8}
\tag{8}
    \mathbf{\Sigma}_{\textbf{n}} = \sum_{i}^{N}\frac{\partial \textbf{n}}{\partial {}^W\textbf{p}_{i}} \mathbf{\Sigma}_{{}^W\textbf{p}_{i}} \frac{\partial \textbf{n}}{\partial {}^W\textbf{p}_{i}} ^{T}
\end{equation}
where the covariance matrix $\mathbf{\Sigma}_{\textbf{n}}$ is the uncertainty of plane 3DOF representation, which calculated from the covariance matrix of the points on the plane $\mathbf{\Sigma}_{{}^W\textbf{p}_{i}}$. 
% Using the Chain rule of function derivation, the expression of $\frac{\partial \textbf{n}}{\partial {}^W\textbf{p}_{i}}$ can be further expressed as \eqref{eq10}.
% \begin{equation}
% \label{eq10}
% \tag{10}
% \frac{\partial \textbf{n}}{\partial {}^W\textbf{p}_{i}} = \frac{1}{\|\textbf{A}\|}\frac{\partial \textbf{A}^{*}}{\partial {}^W\textbf{p}_{i}} - 
% \frac{\textbf{A}^{*}}{\|\textbf{A}\|^2}\frac{\partial \| \textbf{A} \|}{\partial {}^W\textbf{p}_{i}} + \frac{\textbf{A}^{*}}{\|\textbf{A}\|}\frac{\partial \textbf{e}}{\partial {}^W\textbf{p}_{i}}
% \end{equation}

Note that all elements in $\textbf{A}$, $\textbf{A}^{*}$ and $\textbf{e}$ are the summation results, so dynamic programming\cite{Introduction to algorithms} can be easily used to update the 3DOF plane representation incrementally in practice which will effectively reduce the complexity of fitting the plane during update introduced in \ref{section3.3.2:Voxel Map Construction and Update}.

\subsection{Mergeable Voxel Mapping Method}
\subsubsection{Motivation}
\label{section3.3.1:Motivation}
In the mapping method based on the voxels, scholars always ignore the relationship between voxels no matter how the voxel maps are segmented or how small the voxels are. 
It is undeniable that simply treating all voxels as independent individuals is very easy for engineering practice and has strong robustness.
However, if we can effectively identify the relationships between voxels (e.g. whether they belong to the same plane), then using this relationship can reasonably improve the accuracy of voxel maps and further promote the performance of LiDAR(-inertial) odometry.
Meanwhile, merging the voxels with unified properties (such as coplanar) can also effectively save memory usage which is undoubtedly meaningful for commercial low-cost robots.

Based on the above ideas, we have modified the process of construction and update of voxel maps on the basis of previous research \cite{VoxelMap}. Then creatively proposed a voxel merging method based on union-find \cite{union-find}. 
This method can distinguish the plane in different voxels termed as $\mathcal{P}^k$ and merge $\mathcal{P}^f$ to estimate the large plane composed of these small planes together, the large plane termed as $\mathcal{P}^f$. 
These $\mathcal{P}^k$ estimation results are regarded as the measurements with covariance of $\mathcal{P}^f$. 
Afterward, we estimate the plane 3DOF representation and covariance of the $\mathcal{P}^f$ using the least squares method which is significantly more accurate than the estimation results of $\mathcal{P}^k$, thereby improving the accuracy of LiDAR(-inertial) odometry.
\subsubsection{Voxel Map Construction and Update}
\label{section3.3.2:Voxel Map Construction and Update}
Our method is based on spatial hashing to guarantee the real-time performance of the system. 
Precisely, our voxel map is maintained by a hash table, the 3-dimensional space is discretized into each small voxel, the key of the hash table is the identification of each voxel while the value of the hash table is the node of union-find which represent the plane feature in this voxel.

We set the side length of each voxel to 0.5m rather than the default 3.0m in VoxelMap.
It is reasonable because we replace the octo-tree with a node of union-find for plane merging which will be introduced in \ref{Plane Merging Based on Union-Find}. 
Due to the lack of the feature of voxel segmentation, if we want to achieve the same resolution, it is necessary to reduce the side length of voxel.
This mainly affects the memory usage of the hash table.
However, by merging $\mathcal{P}^k$ in different voxels, we can free up the vast majority of voxels to reduce memory usage.
In experiments shown in \ref{resources usage}, we found that the memory usage of VoxelMap++ is even lower than that of VoxelMap since we do not need to maintain the planar representation in each voxel.

After the arrival of the first frame of LiDAR points, we check all of the voxels that contain enough points, then we use PCA\cite{pca} to judge whether the points in each voxel can form a plane. 
After that, we calculate and store the plane parameters $\textbf{n} = [a, b, d]^T$ and their uncertainty $\Sigma_{\textbf{n}}$ based on \eqref{eq6}\eqref{eq8}.
For the LiDAR points that come after the first scan, their poses in the world frame will be obtained through state estimation. 
Then these points are registered into the map. 
If there already exists a voxel in the position of the point and this voxel has not converged, then this point will be added to it and the parameters of the voxel will be updated incrementally based on \eqref{eq6}\eqref{eq8}.
Otherwise, we will construct a new voxel in that position. 

To avoid the growing processing time with the arrival of more lidar points, the update of each voxel will stop when there are more than 50 points in it, because the uncertainty of the parameters of the plane converges when the number of points reaches 50\cite{VoxelMap}. 
When a voxel stops updating, we discard all points in the voxel to save memory and only retain the plane parameters $\textbf{n} = [a, b, d]^T$ and their uncertainty $\Sigma_{\textbf{n}}$. 
Finally, the newly converging set of planes $\mathcal{S}$ will serve as input of the plane merging algorithm in \ref{Plane Merging Based on Union-Find}.

\begin{figure*}[t]
\centering
\subfigure[planes w/o merging]{
\label{Fig.sub.1}
\includegraphics[width=0.31\textwidth]{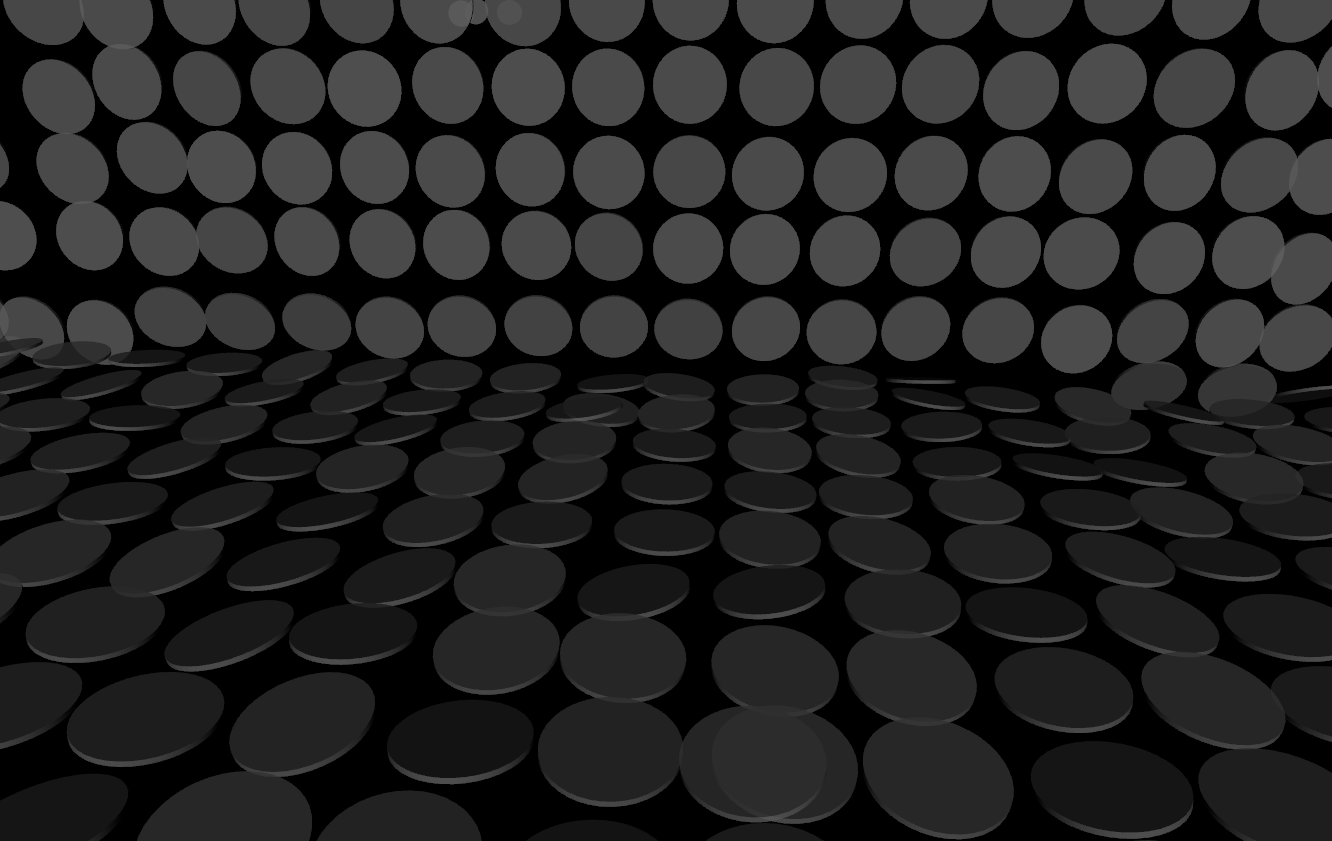}
}\subfigure[planes w merging]{
\label{Fig.sub.1}
\includegraphics[width=0.31\textwidth]{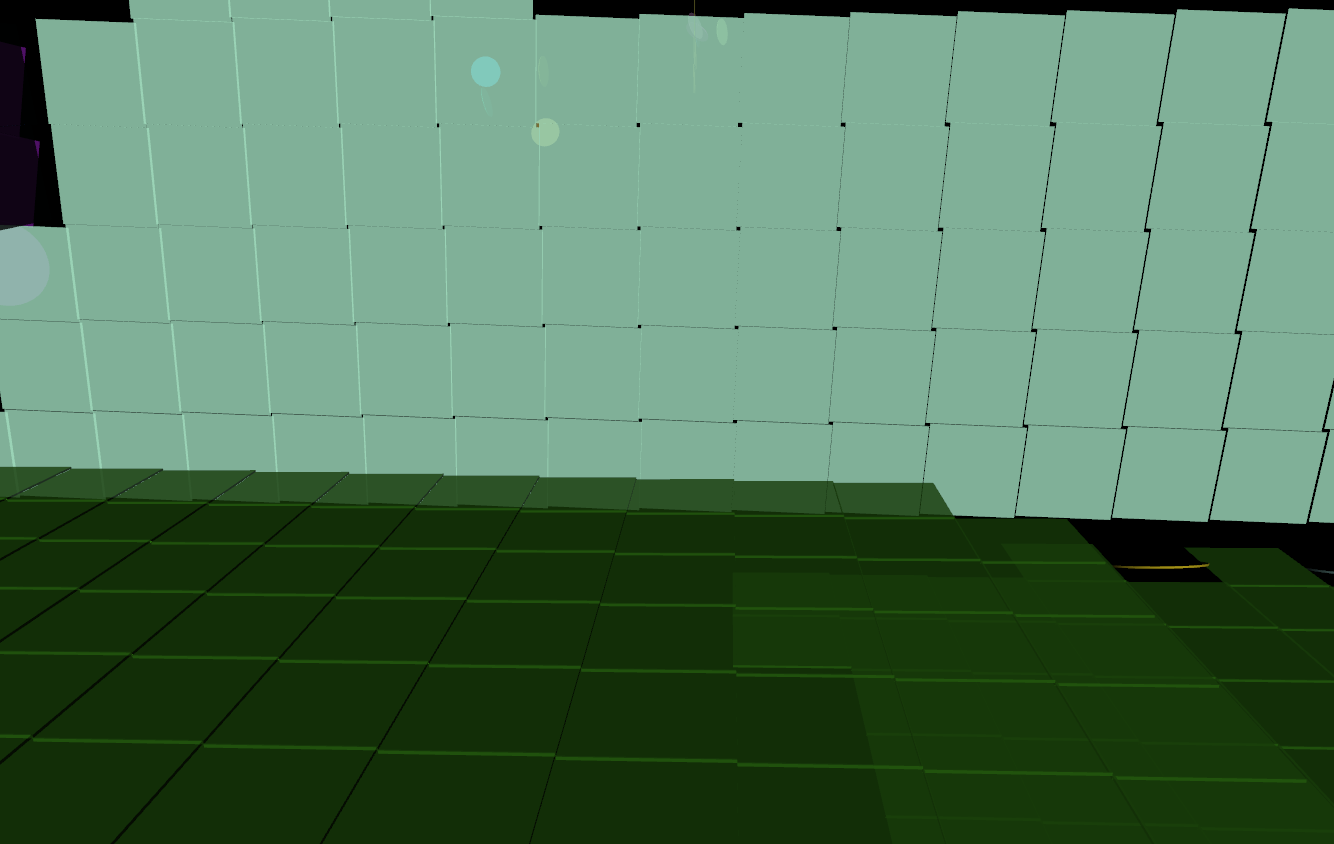}
}\subfigure[reality]{
\label{Fig.sub.2}
\includegraphics[width=0.31\textwidth]{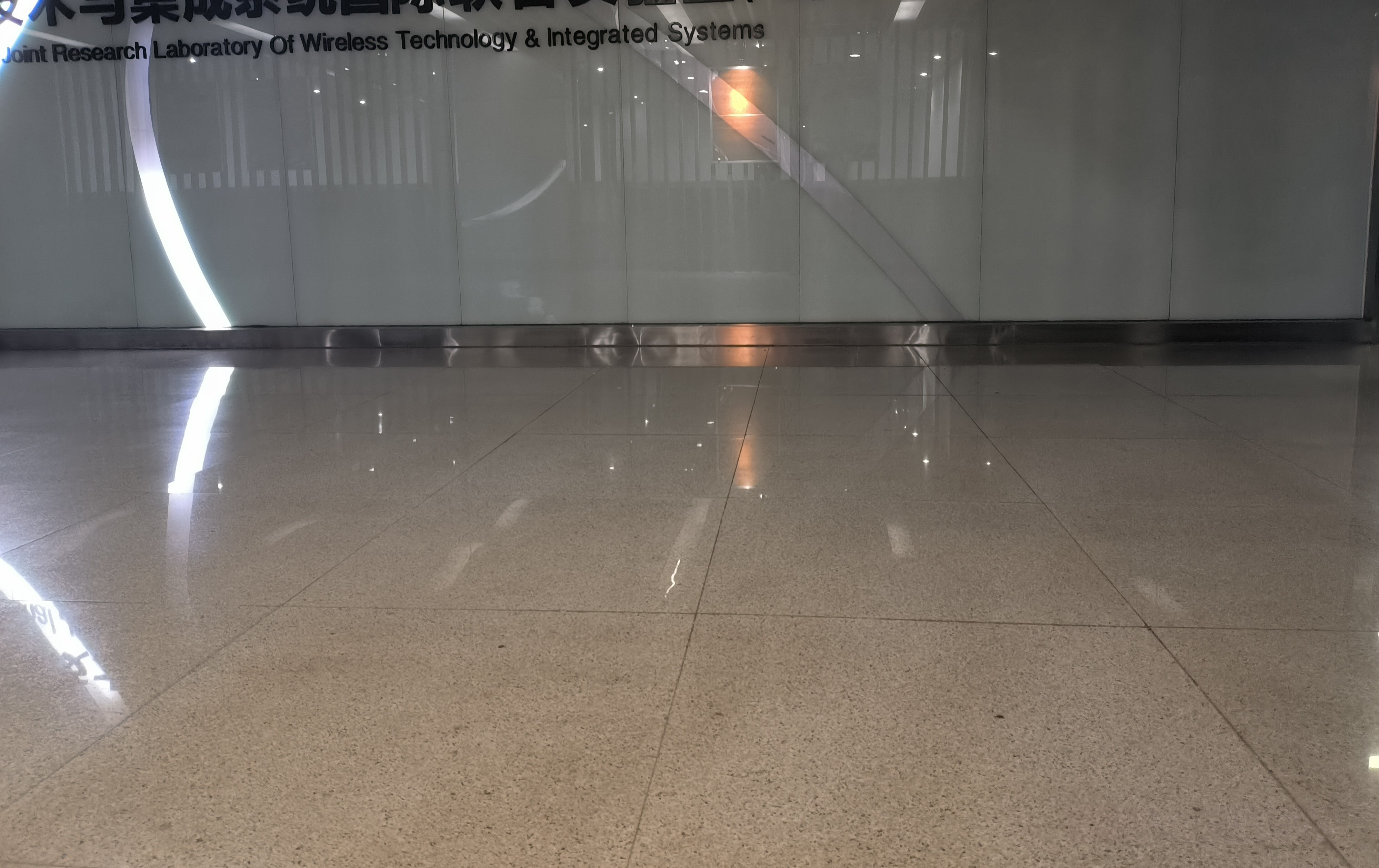}}
\caption{An example of planes w/ and w/o merging. 
In (a), every single plane $\mathcal{P}^k_i$ has its own parameters. 
In (b), the planes with the same color belong to one $\mathcal{P}^f$ and share the plane representation.
In (c), it is the reality scene of background.
Compared with reality, it is obvious that the plane merging method proposed in this subsection can effectively improve the accuracy of plane estimation}
\label{Fig.main}
\end{figure*}
\subsubsection{Plane Merging Based on Union-Find}
\label{Plane Merging Based on Union-Find}
% It is reasonable to merge the planes among neighbor voxels if they represent a same plane in the real physical world such as a large wall or flat ground, it can not only increase the accuracy of plane representation but also reduce the memory consumption.
After voxel map update, converged plane $\mathcal{P}^k$ will be merged with their neighborhood.
The plane merging algorithm is designed based on union-find shown as Algorithm \ref{alg:1}.
\begin{algorithm}
\renewcommand{\algorithmicrequire}{\textbf{Require:}}
\caption{Plane Merging Based on Union-Find}
\label{alg:1}
\begin{algorithmic}[1]
\REQUIRE converging planes set $\mathcal{S}$
    \FOR{$\mathcal{P}^k_i \in \mathcal{S}$}
        \STATE $\mathcal{P}^k_i.f = \mathcal{P}^k_i$
    \ENDFOR 
    \FOR{$\mathcal{P}^k_i \in \mathcal{S}$}
    \FOR{$\mathcal{P}^{n}_{\ } \in \mathcal{P}^k_i.neighbor$}
        \IF {$ \mathcal{P}^{n}_{\ }.f$ is not coplane with $\mathcal{P}^{k}_i.f$}
        \STATE \textbf{continue}
        \ENDIF
        \STATE estimate ${}^{N}\mathcal{P}^k_i.f$ based on \eqref{eq11}\eqref{eq12}
        \IF {$\mathcal{P}^{n}_{\ }.f == \mathcal{P}^{n}_{\ }$ }
        \STATE $\mathcal{P}^{n}_{\ }.f = \mathcal{P}^{k}_i.f$
        \ELSE
        \STATE $Node = max\_kids(\mathcal{P}^{n}_{\ }.f, \mathcal{P}^{k}_i.f)$
        \STATE $\mathcal{P}^{k}_i.f = \mathcal{P}^{n}_{\ }.f = Node$
        \STATE pruning on each kids node about $\mathcal{P}^{k}_i.f$
        \ENDIF
        \STATE $\mathcal{P}^k_i.f = {}^{N}\mathcal{P}^k_i.f$
        \STATE free the memory resource in $\mathcal{P}^k_i$ and $\mathcal{P}^n_{}$
    \ENDFOR  
    \ENDFOR 
\end{algorithmic}  
\end{algorithm}

\begin{figure}[htb]
\centering
\includegraphics[width=0.49\textwidth]{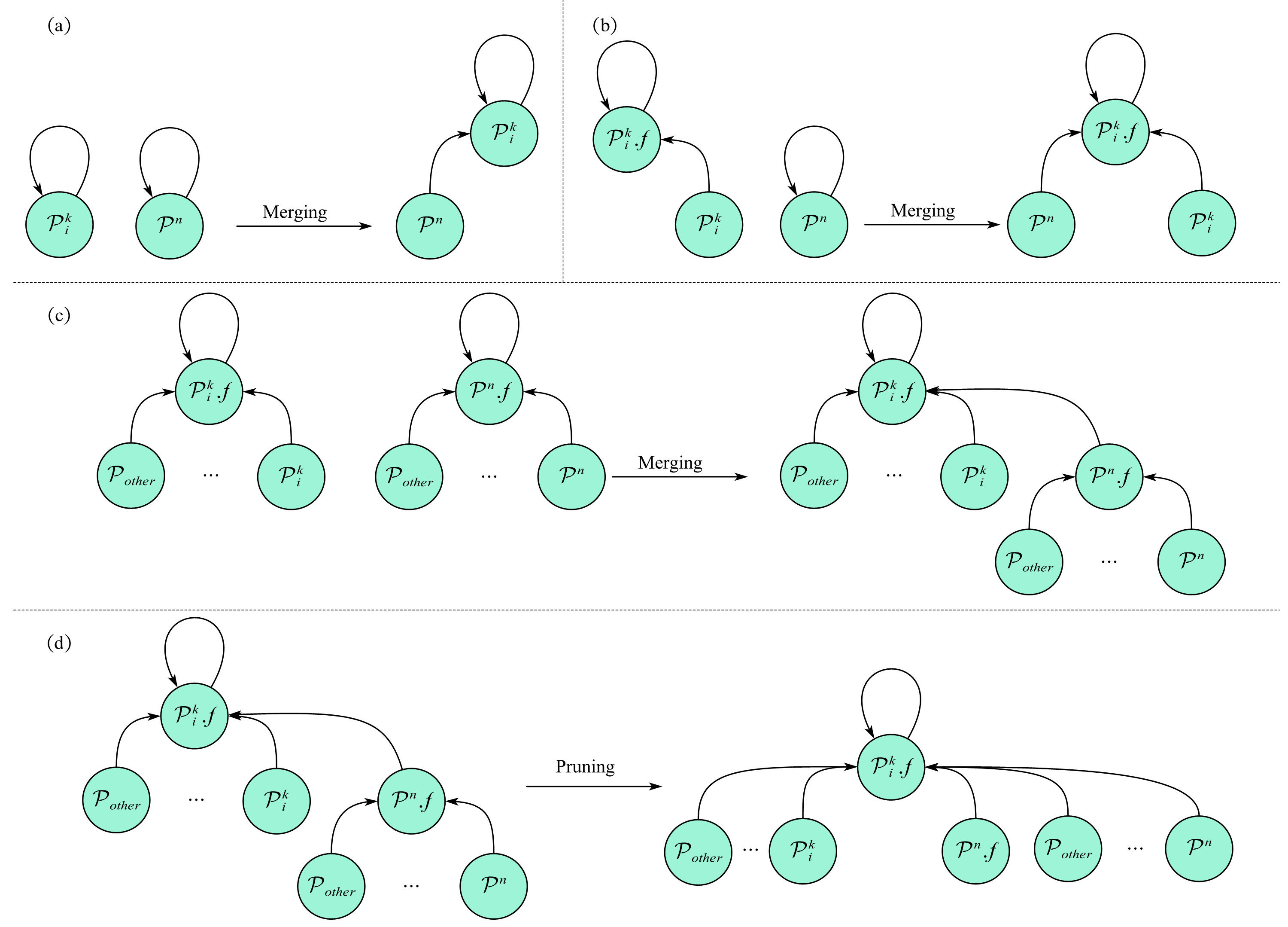}
\caption{The Plane Merging and Node pruning method in Algorithm 1}
\label{Fig.MergingAndPruning}
\end{figure}

In 1-3, the new node in union-find which is the converged plane will be initialized. 
Then in 4-17, these new nodes will merge with other nodes in the union-find.
In 5-16, each neighbor plane of $\mathcal{P}^{k}_i$ has been queried based on hash key and performing the plane merging about this $\mathcal{P}^{k}_i$ and $\mathcal{P}^{n}_{\ }$.
In 6-8, we will calculate the similarity of $\mathcal{P}^{k}_i$ with $\mathcal{P}^{n}_{\ }$ based on the Mahalanobis distance shown in \eqref{eq10}.
These two planes will be merged when the Mahalanobis distance is less than the threshold given by the 95\% of the $\chi ^2$ distribution. 
In 9, we regard the $\mathcal{P}^k_i.f$ and $\mathcal{P}^n.f$ as the measurement with covariance of the larger plane ${}^N\mathcal{P}^k_i.f$, and estimate it covariance and 3DOF representation according \eqref{eq11}\eqref{eq12} which is a simple weighted average algorithm based on minimum the trace of $\mathbf{\Sigma}_{\textbf{n}_{{}^{N}\mathcal{P}^{k}_i}}$.  
In 10-12, it is two simple cases of merging nodes based on union-find which is shown as Fig.\ref{Fig.MergingAndPruning}(a,b).
In 13-15, it is another complex case about plane merging about $\mathcal{P}^k_i.f$ and $\mathcal{P}^n.f$ Fig.\ref{Fig.MergingAndPruning}(c). 
In 16, we perform pruning after merging to maintain the maximum depth of union-find is 2 to avoid additional time consumption which is shown in Fig.\ref{Fig.MergingAndPruning}(d).
In 17, we assign the results of the larger plane $^{N}\mathcal{P}^k_i.f$ to $\mathcal{P}^k_i.f$ , then $\mathcal{P}^{n}_{\ }.f$ and $\mathcal{P}^{k}_i.f$ have been merged successfully. 
Finaly, in 18, we free the resource which is the 3DOF representation and $3\times3$ covariance in $\mathcal{P}^k_i$ and $\mathcal{P}^n_{}$.
After merging, hundreds of voxels will share the same plane representation in $\mathcal{P}^k.f$,which will effectively reduce the memory usage.
\begin{gather}
\label{eq10}
\tag{9}
\gamma = (\textbf{n}_{\mathcal{P}^{k}_i} - \textbf{n}_{\mathcal{P}^{n}_{\ }})(\mathbf{\Sigma}_{\textbf{n}_{\mathcal{P}^{k}_i}} + \mathbf{\Sigma}_{\textbf{n}_{\mathcal{P}^{n}_{\ }}})^{-1}(\textbf{n}_{\mathcal{P}^{k}_i} - \textbf{n}_{\mathcal{P}^{n}_{\ }})^{T}\\
\label{eq11}
\tag{10}
 \textbf{n}_{{}^{N}\mathcal{P}^{k}_i} = \frac{trace(\mathbf{\Sigma}_{\textbf{n}_{\mathcal{P}^{n}_{\ }}}) \textbf{n}_{\mathcal{P}^{k}_i} + trace(\mathbf{\Sigma}_{\textbf{n}_{\mathcal{P}^{k}_{i}}}) \textbf{n}_{\mathcal{P}^{n}_{\ }}}{trace(\mathbf{\Sigma}_{\textbf{n}_{\mathcal{P}^{k}_{i}}}) + trace(\mathbf{\Sigma}_{\textbf{n}_{\mathcal{P}^{n}_{\ }}}) } \\
\label{eq12}
\tag{11}
\mathbf{\Sigma}_{\textbf{n}_{{}^{N}\mathcal{P}^{k}_i}}  = \frac{trace(\mathbf{\Sigma}_{\textbf{n}_{\mathcal{P}^{n}_{\ }}})^2\mathbf{\Sigma}_{\textbf{n}_{\mathcal{P}^{k}_{i}}} + trace(\mathbf{\Sigma}_{\textbf{n}_{\mathcal{P}^{k}_{i}}})^2\mathbf{\Sigma}_{\textbf{n}_{\mathcal{P}^{n}_{\ }}}}{\big(trace(\mathbf{\Sigma}_{\textbf{n}_{\mathcal{P}^{k}_{i}}}) + trace(\mathbf{\Sigma}_{\textbf{n}_{\mathcal{P}^{n}_{\ }}})\big)^2 }   
\end{gather}

Fig.\ref{Fig.sub.1} is the figure of planes without merging, and Fig.\ref{Fig.sub.2} is the planes after merging, planes with the same color belong to one plane and share the fitting results $\textbf{n}$ and covariance $\Sigma_{\textbf{n}}$.
% Merge process only happens between adjacent planes. The similarity between planes is judged by the Mahalanobis distance instead of Euclidean distance because Mahalanobis distance will take the covariance of parameters into consideration. Adjacent planes will be merged when the Mahalanobis distance is less than the threshold given by the 95\% of the $\chi ^2$ distribution. The Mahalanobis distance $\gamma$ between planes is:
% \begin{equation}
%     \gamma = \textbf{r}(\mathbf{\Sigma}_{\textbf{n}_1} + \mathbf{\Sigma}_{\textbf{n}_2})^{-1}\textbf{r}^{T}
% \end{equation}
% where,
% \begin{equation}
%     \textbf{r} = \textbf{n}_1 - \textbf{n}_2
% \end{equation}

% When two planes satisfy the merge condition mentioned above, the next step we take a strategic approach to computing the combined plane parameters. Let $[a_{m}, b_{m}, d_{m}]^{T}$ be the parameters of the plane after merging, then we get:
% \begin{align}
%     & \left[ \begin{matrix} a_m \\\ b_m \\\ d_m \end{matrix} \right] = \frac{\left\|\mathbf{\Sigma}_{a_2,b_2,c_2}\right\|_F \left[ \begin{matrix} a_1 \\\ b_1 \\\ d_1 \end{matrix} \right] + \left\|\mathbf{\Sigma}_{a_1,b_1,c_1}\right\|_F \left[ \begin{matrix} a_2 \\\ b_2 \\\ d_2 \end{matrix} \right]}{\left\|\mathbf{\Sigma}_{a_1,b_1,c_1}\right\|_F + \left\|\mathbf{\Sigma}_{a_2,b_2,c_2}\right\|_F }
% \end{align}
% and $\mathbf{\Sigma}_{a_m,b_m,d_m}$ be its covariance(11):
\subsection{State Estimation based on IESKF}
The state estimation is based on an
iterated extended Kalman filter similar to FAST-LIO\cite{FAST-LIO} and VoxelMap\cite{VoxelMap}. 
Assume that we are given a state estimation prior $\hat{\textbf{x}}_k$ with covariance $\hat{P}_k$ based on IMU propagation.
This prior will be updated with the point-to-plane distance matched to form a maximum a posteriori (MAP) estimation. Specifically, the i-th valid point-to-plane match leads to the observation equation shown as
\begin{align}
\label{eq13}
\tag{12}
   z_i & = h_i(\textbf{x}_i, \textbf{n}_i) = 0\\
   & 
    \notag
    \approx h_i(\hat{\textbf{x}}^\kappa_i, 0) + \textbf{H}_i^\kappa(\textbf{x}_i \boxminus \hat{\textbf{x}}_i^\kappa) + \textbf{v}_i  \\
    \label{eq13}
    \tag{13}
& h_i(\hat{\textbf{x}}^\kappa_i, 0) = \frac{\Omega^T({}_{L}^{W}\textbf{R}{}^L\textbf{p}_i + {}^W_L\textbf{t}) + d}{||\Omega||}   
\end{align}
where $h_i(\hat{\textbf{x}}^\kappa_i, 0)$ is the point-to-plane distance in the $\kappa$ iteration and $\Omega$ is the normal vector $[a,b,1]^T$ of registration plane $\mathcal{P}^{k}_i.f$. 
$\mathcal{P}^{k}_i.f$ can be determined by spatial hash querying $\mathcal{P}^{k}$ from voxelmap firstly, then finding the root nodes of  $\mathcal{P}^{k}$ based on union-find.
The $\textbf{v}_j \sim (0, \textbf{R}_j)$ is the observation noise propagate from the ${}^L\textbf{p}_i$ and $\textbf{n}_{\mathcal{P}^{k}_i.f}$ which is shown as
\eqref{eq14}.
\begin{gather}
    \notag
    \Sigma_{\textbf{n}_{\mathcal{P}^{k}_i.f}, {}^L\textbf{p}_i}= \left[\begin{matrix}
        \Sigma_{\textbf{n}_{\mathcal{P}^{k}_i.f}}&  \textbf{0}_{3\times3} \\
        \textbf{0}_{3\times3} & \Sigma_{{}^L\textbf{p}_i}
    \end{matrix}\right]\\
    \notag
    \textbf{R}_i = \textbf{J}_{\textbf{v}_i}\Sigma_{\textbf{n}_{\mathcal{P}^{k}_i.f}, {}^L\textbf{p}_i}\textbf{J}_{\textbf{v}_i}^T \\
    \tag{14}
    \label{eq14}
    \textbf{J}_{\textbf{v}_i} = [\textbf{J}_{\textbf{n}_i}, \textbf{J}_{{}^L\textbf{p}_i}], \textbf{J}_{{}^L\textbf{p}_i} = \frac{1}{||\Omega||}\Omega^T\times{}^W_L\textbf{R} \\
    \notag
    \textbf{J}_{\textbf{n}_i} = \frac{1}{||\Omega||} \left[\begin{matrix}
        x_i(1  - \frac{1}{||\Omega||^2}h_i(\textbf{x}_i, 0)) \\
        y_i(1 - \frac{1}{||\Omega||^2}h_i(\textbf{x}_i, 0)) \\
        1
    \end{matrix}\right]^T
\end{gather}

Finally, combining the state prior with all effective measurements, we can obtain the MAP estimation:
\begin{gather}
\label{eq16}
\tag{15}
\min_{\hat{\textbf{x}}_k^\kappa}
 \left\{  ||\hat{\textbf{x}}_k^\kappa \boxminus \hat{\textbf{x}}_k||^2_{\hat{\textbf{P}}_k} + \sum_{i = 1}^{N}|| h_i(\hat{\textbf{x}}^\kappa_k, 0) + \textbf{H}_i^\kappa(\textbf{x}_k \boxminus \hat{\textbf{x}}^\kappa_k )||_{\textbf{R}_i} \right\}
\end{gather}
where the first part is the state prior and the second part is the
measurement observation. The detail solution is based on an iterated extended
Kalman filter can refer \cite{IESKFORM}.
\section{Experiments}
\label{section2:Replated Work}
We implemented the proposed VoxelMap++ system in C++
and Robots Operating System (ROS) on a laptop computer with 2.9GHz 8 cores and 16Gib memory.
The experiment data includes open source datasets M2DGR\cite{M2DGR} and our own challenging degenerated or unstructured datasets. 

The M2DGR dataset used a Velodyne VLP-32C with $360^\circ \times 40^\circ$ FOV to scan the surrounding environment and obtain the 3D point cloud in 10Hz.
It also used a VI-sensor Realsense d435i to obtain inertial data at 200Hz.

\begin{figure}[htb]
\centering
\includegraphics[width=0.45\textwidth]{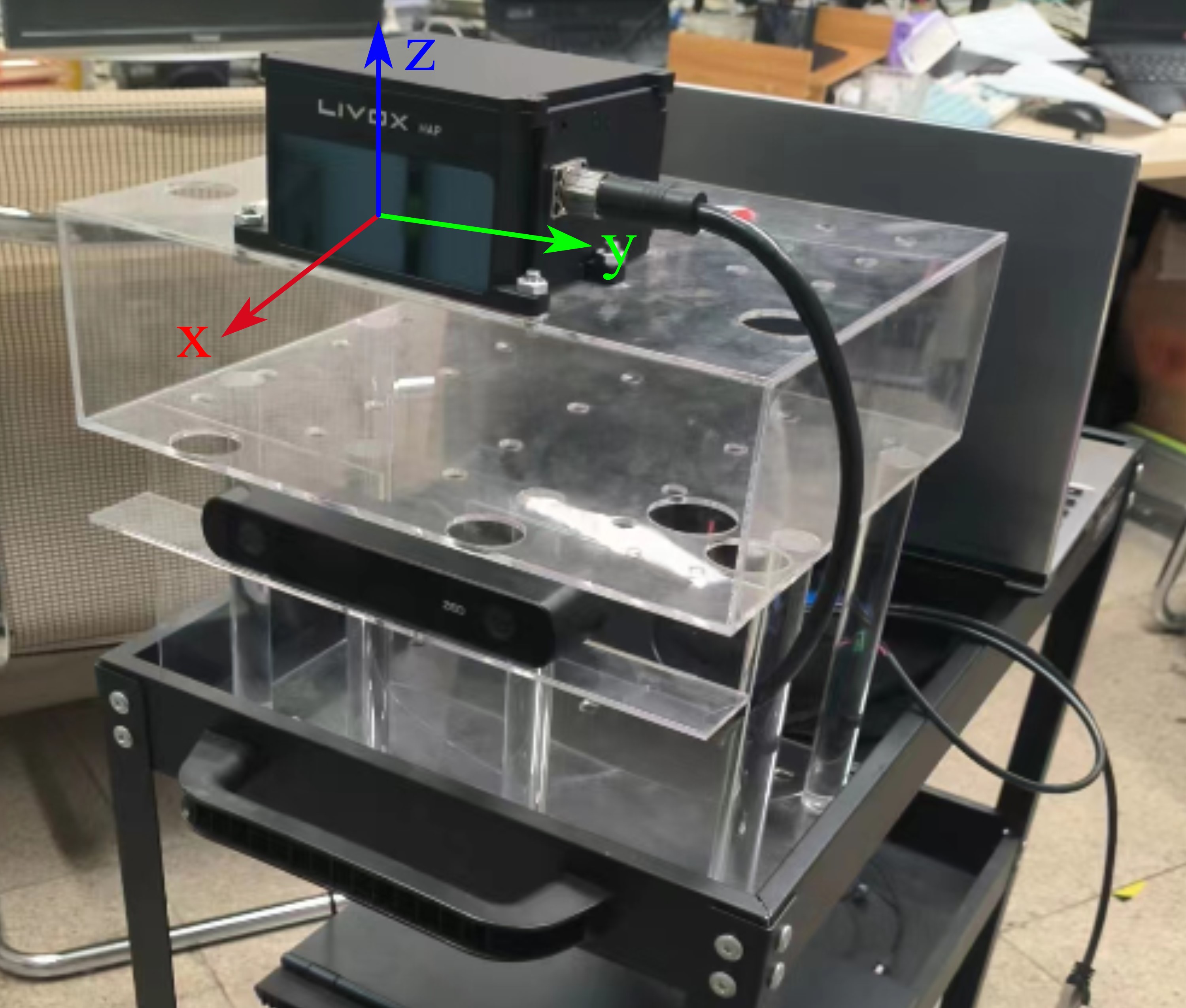}
\caption{Our data collection equipment has Livox HAP LiDAR and ZED 2i camera with inner IMU, these devices are well strapdown on the trolley}
\label{Fig.data platform}
\end{figure}

Our own dataset is collected via a solid-state Livox HAP LiDAR with $120^\circ \times 25^\circ$ FOV which is the first automotive-grade LiDAR for serial production, and a ZED 2i camera with built-in Next-Gen IMU with gyroscope, accelerator, barometer, and magnetometer.
The frequency of LiDAR point cloud and IMU is 10Hz and 500Hz, respectively.
These two sensors are synchronized based on IEEE 1588-2008 and the extrinsic of LiDAR and IMU ${}_{L}^{W}\textbf{R}, {}_{L}^{W}\textbf{t}$ have been calibrated by using LI-Init\cite{LI-Init}.
The sensors platform is shown as Fig.\ref{Fig.data platform}. 
Our own datasets will also be open source to benefit researchers.

\subsection{Experiment on Structured Urban}
M2DGR covers the environment of structured urban, this dataset is collected by a ground robot with a full sensor-suite including an inertial measurement unit and a LiDAR, while it is also equipped with a GNSS-IMU system with real-time kinematic signals to get the dataset's ground truth. All those sensors are well-calibrated and synchronized.

% The M2DGR has already provided some results of the state-of-the-art methods such as A-LOAM, LeGO-LOAM, LINS and LIO-SAM, so we will utilize their results and add the results of VoxelMap and our method.
% \begin{table*}
%     \caption{SAMPLE SEQUENCES FOR EVALUATION}
%     \centering
%     \begin{tabular}{cccccccc}
%     \hline
%     Sequence & Street02 & Street06 & Street07 & Roomdark06 & Hall05 & Door01 & Lift04 \\
%     \hline
%     Duration/s & 1227 & 494 & 929 & 172 & 402 & 461 & 299 \\
%     Distance/m & 1484.62 & 479.63 & 1104.07 & 72.53 & 79.28 & 285.51 & 142.78 \\
%     Speed/(m/s) & 1.21 & 0.97 & 1.19 & 0.42 & 0.71 & 0.31 & 0.27 \\
%     \hline
%     Description & day, & night, & nigh,t & room, & long-term, & outdoors to & first floor to second \\
%     of features & long-term & straight line & zigzag route & complete darkness & large overlap & indoors & floor by lift \\
%     \hline
%   \end{tabular} 
% \end{table*}
\begin{table*}[tb]
    \caption{accuracy (ate in meters) comparison on m2dgr structured urban sequences}
    \centering
    \begin{tabular}{cccccccc}
    \hline
    Sequence & Street02(a) & Street06(b) & Street07(c) & Roomdark06(d) & Hall05(e) & Door01(f) & Lift04(g) \\
    \hline
    A-LOAM & 5.299 & 0.628 & 28.940 & 0.314 & 1.065 & 0.274 & 1.323 \\
    LeGO-LOAM & 20.021 & 1.246 & 35.437 & 0.373 & 1.030 & 0.253 & 1.370 \\
    LIO-SAM & 4.063 & 0.417 & 28.642 & 0.324 & 1.047 & 0.268 & X \\
    LINS & 5.636 & 1.742 & 12.009 & 2.205 & 1.010 & 0.258 & \textbf{1.318} \\
    FAST-LIO2 & 2.3236 & 0.4570 & 11.7518 & 0.3146 & 1.0227 & 0.2566 & X \\
    Faster-LIO & 2.6667 & 0.4138 & \textbf{11.7363} & 0.3124 & 1.0311 & 0.2522 & X \\
    VoxelMap & 1.7408 & 0.4901 & 13.7607 & 0.2944 & 0.9376 & 0.2361 & X \\
    VoxelMap++ & \textbf{1.1608} & \textbf{0.4118} & 12.8530 & \textbf{0.2533} & \textbf{0.8991} & \textbf{0.2170} & X \\
    \hline
    \hline
    Duration/s & 1227 & 494 & 929 & 172 & 402 & 461 & 299 \\
    Distance/m & 1484.62 & 479.63 & 1104.07 & 72.53 & 79.28 & 285.51 & 142.78 \\
    Speed/(m/s) & 1.21 & 0.97 & 1.19 & 0.42 & 0.71 & 0.31 & 0.27 \\
    Description & day, & night, & night & room, & long-term, & outdoors to & first floor to second \\
    of features & long-term & straight line & zigzag route & complete darkness & large overlap & indoors & floor by lift \\
    \hline
    \hline
    \label{table 1}
  \end{tabular} 
\end{table*}

Table \ref{table 1} gives detailed information about tests route and evaluation results on A-LOAM, LeGO-LOAM, LIO-SAM, LINS, FAST-LIO2, VoxelMap and our proposed VoxelMap++. 
It is clear that these sequences include various environments for SLAM including long distance and short distance, indoors and outdoors, straight line and zigzag route. 
These scenarios are sufficient to illustrate the structured urban environment.

We use the absolute trajectory error (ATE)\cite{ATE} to illuminate the accuracy, where the results
for A-LOAM, LeGO-LOAM, LIO-SAM and LINS are directly drawn from \cite{M2DGR}.
ATE of FAST-LIO2, VoxelMap and VoxelMap++ is calculated based on evo\cite{evo} which is  an open-source trajectory error evaluation toolkit.
It is obviously that our method VoxelMap++ has better accuracy than other algorithms including VoxelMap under most circumstances. 
It should be noted that for sequence lift04, the positioning results of VoxelMap and our method fail from a position.
This is because the lift04 sequence includes rising in a closed elevator.
Unfortunately, the voxel on the elevator gates has converged before closed which means the map will not change anymore on VoxelMap and VoxelMap++.
When the elevator gates closed, the positioning results diverges rapidly which illuminate that VoxelMap and our proposed algorithm not applicable in dynamic scenes.

The reason our method is more accurate than VoxelMap and other state-of-art algorithm is that we adopt plane merging block which mean more points will be used to describe a single plane. 
In point cloud based methods such as FAST-LIO2, only 5 points are used to fit the plane commonly because of the limitation of the real-time performance of the system, which makes the estimation of the plane not accurate enough for positioning.
In VoxelMap, points within a voxel are used for plane fitting, but ignore the planes relationship between voxels.
Therefore, its accuracy improvement is also limited.
In VoxelMap++, our plane merging block based on union-find \ref{Plane Merging Based on Union-Find} can better estimate the plane fitting result and covariance by making good use of the coplanar information about entire voxel map.
In some cases, the laser points on the entire wall or the entire floor can be used to calculate the parameters of the single huge plane such as ground.

\begin{figure}[htb]
\centering
\subfigure[]{
\label{Fig.sub.3}
\includegraphics[width=0.14\textwidth]{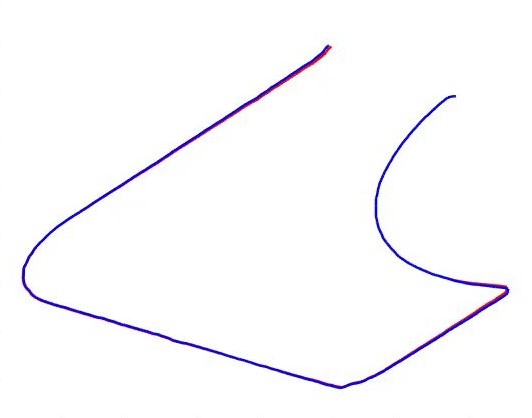}}
\subfigure[]{
\label{Fig.sub.4}
\includegraphics[width=0.14\textwidth]{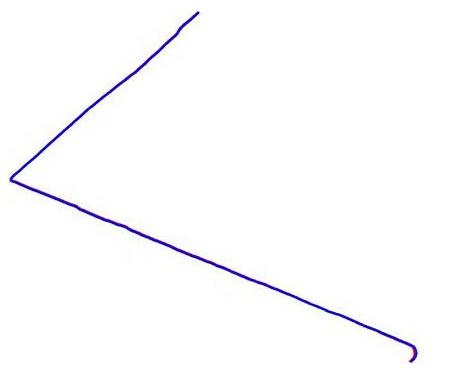}}
\subfigure[]{
\label{Fig.sub.3}
\includegraphics[width=0.14\textwidth]{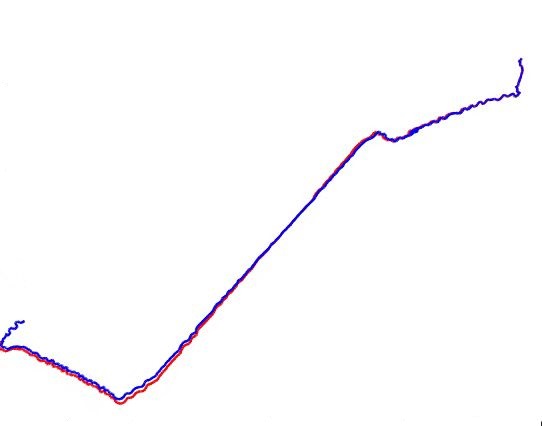}
}
\subfigure[]{
\label{Fig.sub.4}
\includegraphics[width=0.14\textwidth]{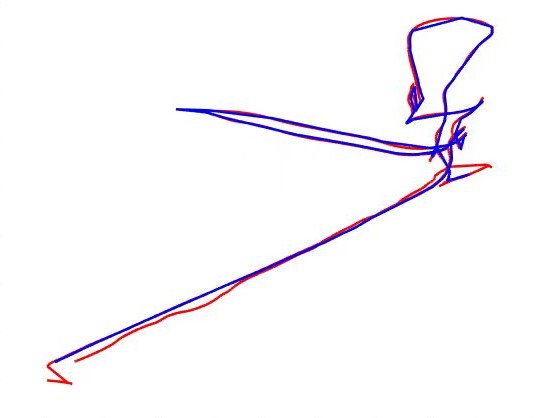}}
\subfigure[]{
\label{Fig.sub.3}
\includegraphics[width=0.14\textwidth]{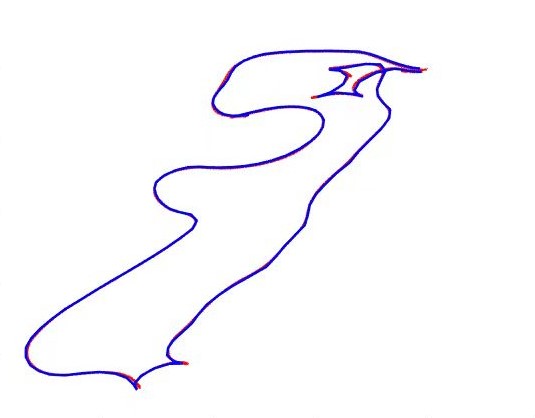}}
\subfigure[]{
\label{Fig.sub.4}
\includegraphics[width=0.14\textwidth]{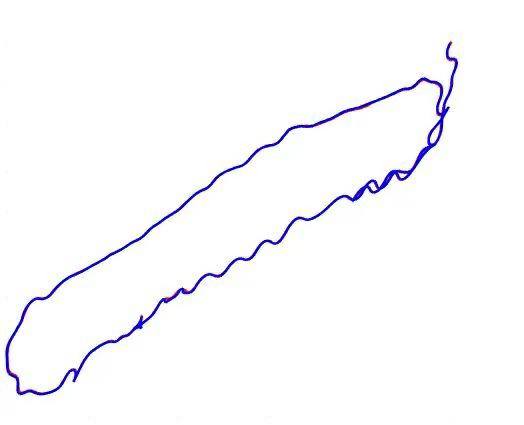}}
\caption{(a)-(f) shows the estimated LiDAR trajectories of our method (blue) and their ground-truth (red) on all sample sequences except sequence Lift04.}
\label{Fig.traj}
\end{figure}

Fig.\ref{Fig.traj} shows the LiDAR trajectories of our method and their ground truth on all sample sequences. 
It is obvious that the trajectories of our method are very close to the ground truth, which means our method has high accuracy.  
Notice that we use the same parameters in all sequences experiment rather than deliberate tuning the parameter to show better results.

\subsection{Experiment on Challenging scenarios}
In order to further test the performance of our algorithm in other challenging scenarios, such as unstructured scenes or indoors with long corridors.
We design a series of experiments on these scenarios based on our own data collection platform.
Unfortunately, due to the lack of hardware for GNSS-IMU systems with real-time kinematic signals.
We can only use end-to-end errors, which are defined as the difference between the start point and terminal point, to compare the accuracy of different algorithms.
In order to estimate the end-to-end error, we will always push the cart with the data collection platform in the path around challenging scenarios and return to the origin finally.
Because the LiDAR in the platform is the non-repetitive scanning Livox HAP, so we can only use the state-of-art algorithm which is applicable for this sensor for comparison, such as FAST-LIO2, Faster-LIO, VoxelMap, and Livox's official LIO-Livox. 
\subsubsection{Unstructured Forests and Grassland}
\begin{figure}[htb]
\centering
\subfigure[]{
\label{outdoor1}
\includegraphics[width=0.22\textwidth]{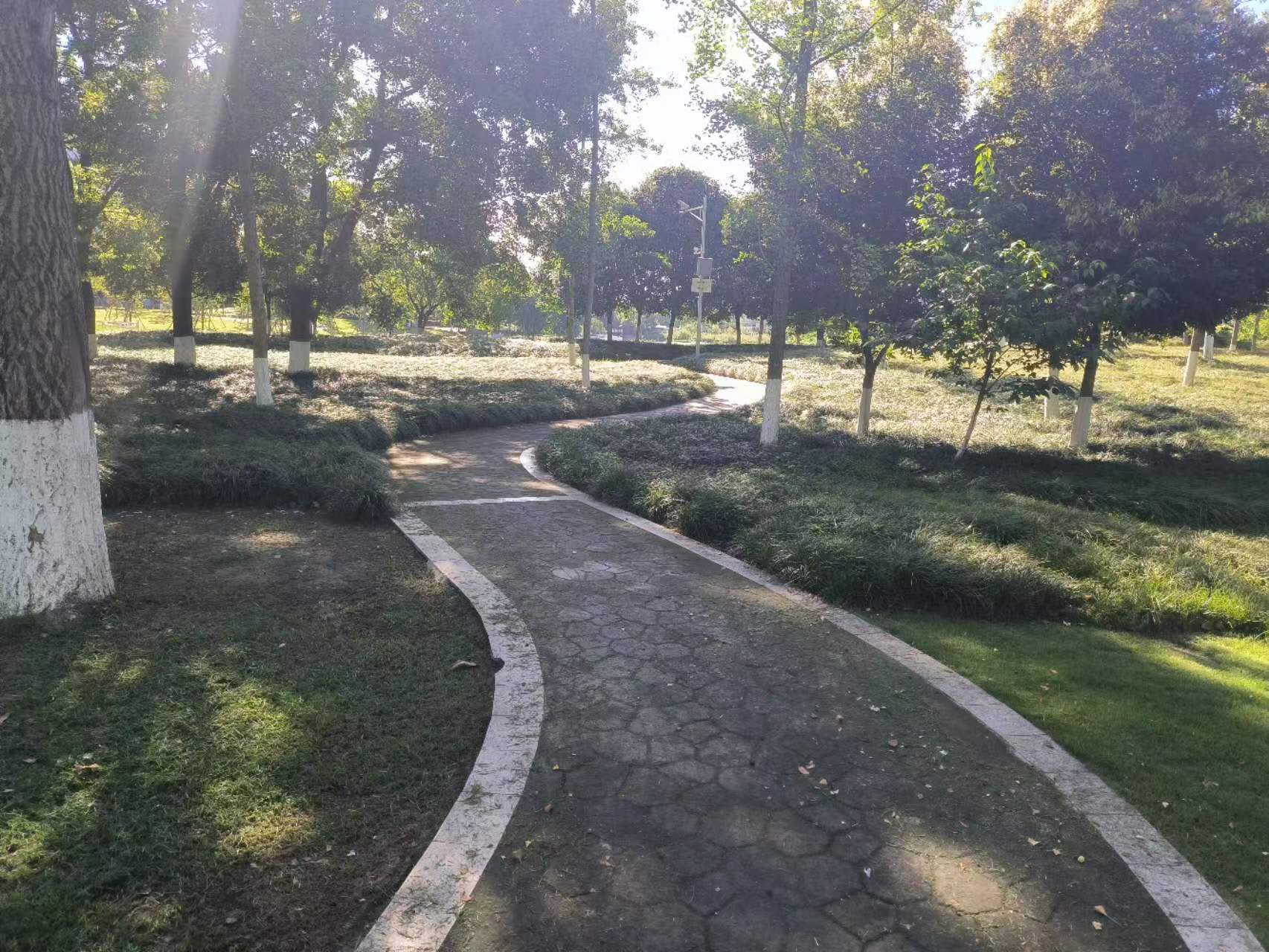}
}\hspace{-0.2cm}
\subfigure[]{
\label{outdoor3}
\includegraphics[width=0.22\textwidth]{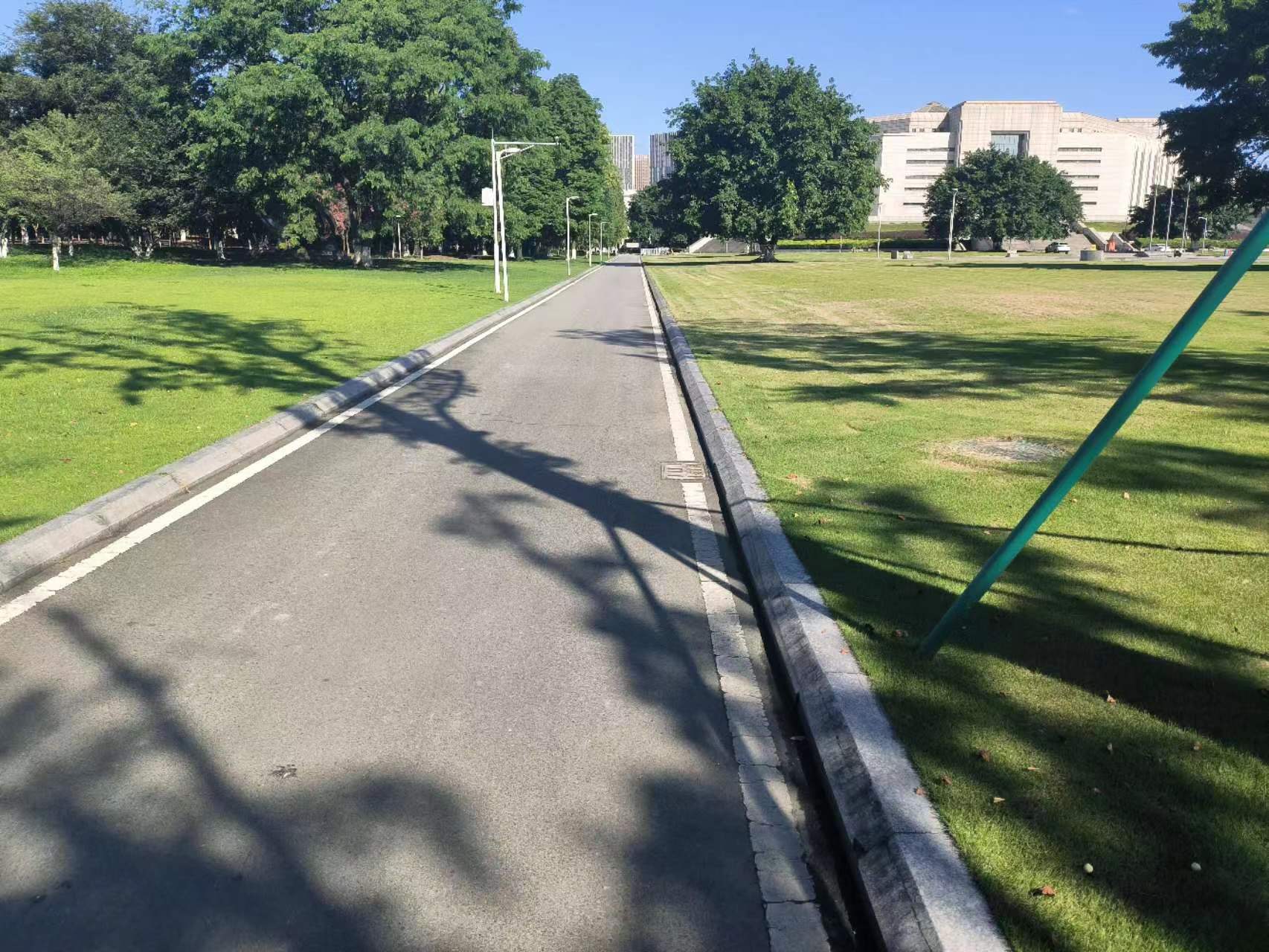}}
\caption{(a)(b) is the outdoor unstructured scenarios of our experiment.}
\label{unstructured scenarios}
\end{figure}

Fig. \ref{unstructured scenarios} shows the unstructured scenarios of our experiment.
we conduct experiments in a forest and grassland at the  entrance of the library in UESTC.

As shown in Table.\ref{table unstructured sequences}, both VoxelMap and VoxelMap++ are more robust and accurate than other start-of-art methods in unstructured scenarios.
This promotion is mainly due to the covariance estimation method in the point cloud $\Sigma_{{}^W\textbf{p}_i}$, plane features $\Sigma_{\textbf{n}}$ and observation noise $\textbf{R}$.
Our proposed VoxelMap++ is more accurate than VoxelMap because the plane merging module can make the estimation of the plane more accurate and the covariance decline in the huge plane. 
This means that the IESKF tends to rely on large planes with low observation covariance rather than chaotic features in forests and grasslands.

Other methods such as FAST-LIO2 regard the noise in each point-to-plane distance as a constant related to the accuracy of sensors, 
which means the features in the messy objects such as tree leaf and weed have the same influence as planar points.
Actually, these features in messy objects do not satisfy the assumption of point in plane any more.
Therefore, these features should be directly deleted or adjust their covariance scientifically. 
The covariance propagation in VoxelMap and VoxelMap++ is the one useful strategy.

\begin{table}[htb]
\caption{end-to-end error(meters) on unstructured sequences}
\centering
\begin{tabular}{cccccccc}
\hline
Sequence & loopE & loopF & loopG & loopH & loopI \\
\hline
LIO-Livox & 5.9141 & 2.8160 & 3.6342 & 2.2157 & 1.2061 \\
FAST-LIO2 & 3.2134 & 0.0830 & 4.2893 & 0.8175 & 0.1591\\
Faster-LIO & 5.7331 & 1.3926 & 1.6632 & 1.7050 & 0.4856\\
VoxelMap & 1.6847 & 0.9577 & 0.4433 & \textbf{0.0492} & 0.0894 \\
VoxelMap++ & \textbf{0.0336} & \textbf{0.0441} & \textbf{0.0389} & 0.0734 & \textbf{0.0406} \\
\hline
Route Length(m) & 329.4 & 373.6 & 311.8 & 519.6 & 284.4 &   \\
\hline
\label{table unstructured sequences}
\end{tabular} 
\end{table}

\subsubsection{Indoor Degenerated Corridor}
\begin{figure}[htb]
\centering
\subfigure[]{
\label{Fig.sub.11}
\includegraphics[width=0.22\textwidth]{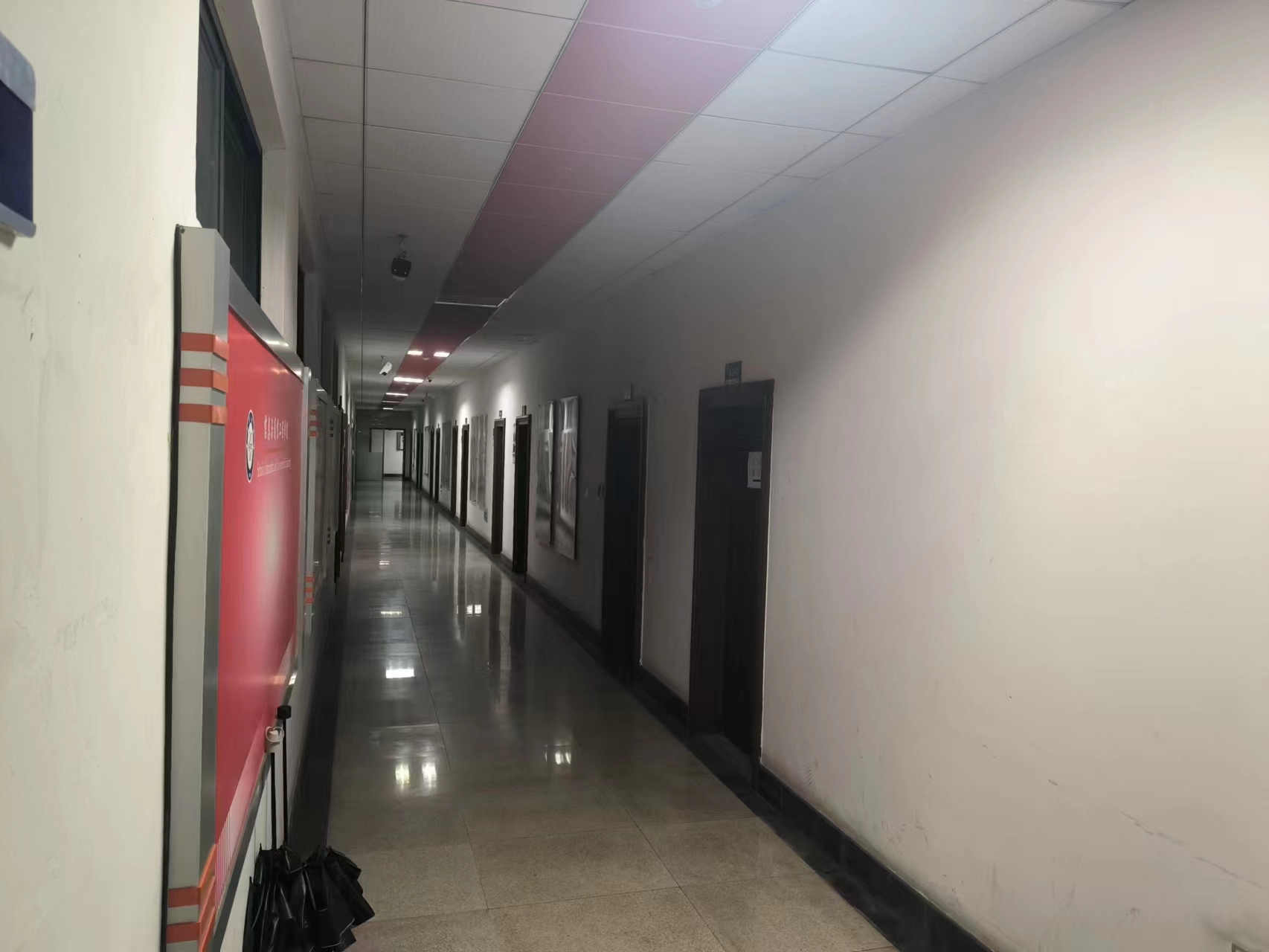}
}\hspace{-0.2cm}
\subfigure[]{
\label{Fig.sub.12}
\includegraphics[width=0.22\textwidth]{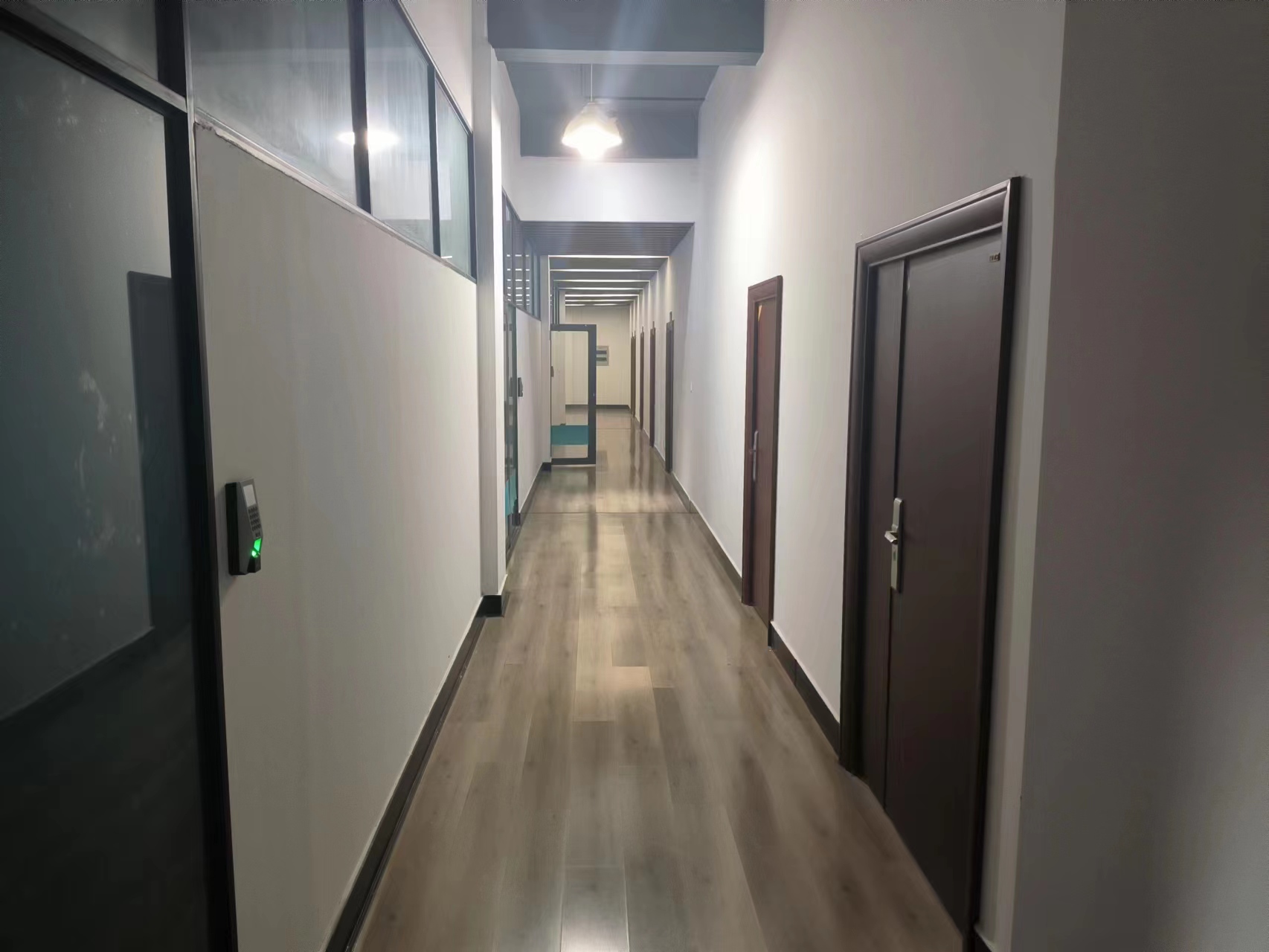}}
\caption{(a)(b) is the indoor degenerated scenarios of our experiment.}
\label{indoor corridor} 
\end{figure}

Fig. \ref{indoor corridor} shows the indoor degenerated scenarios of our experiment.
We conduct experiments inside a building with a lot of long corridors to evaluate the performance of VoxelMap++ in  indoor challenging scenarios.

\begin{figure}[htb]
\centering
\subfigure[]{
\label{Fig.sub.13}
\includegraphics[width=0.22\textwidth]{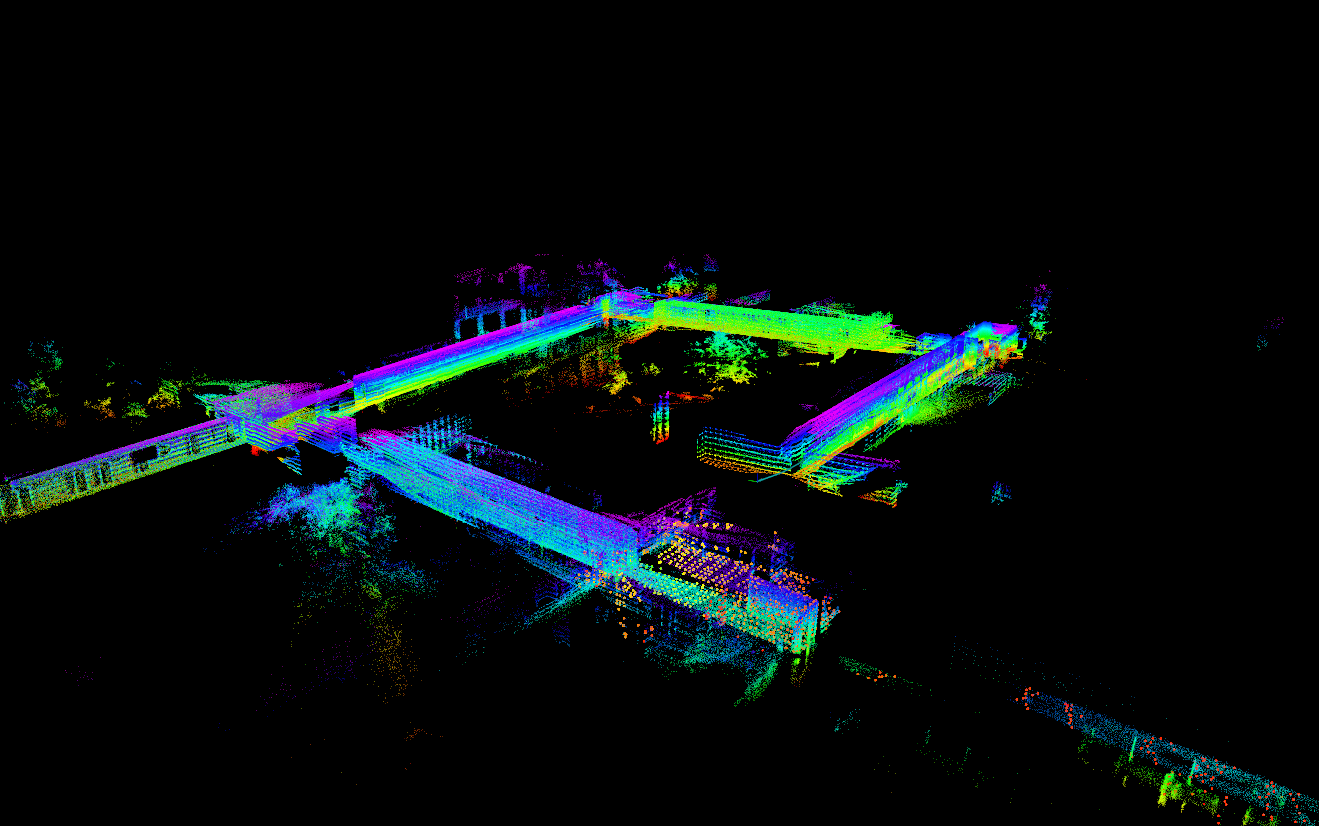}
}\hspace{-0.2cm}
\subfigure[]{
\label{Fig.sub.14}
\includegraphics[width=0.22\textwidth]{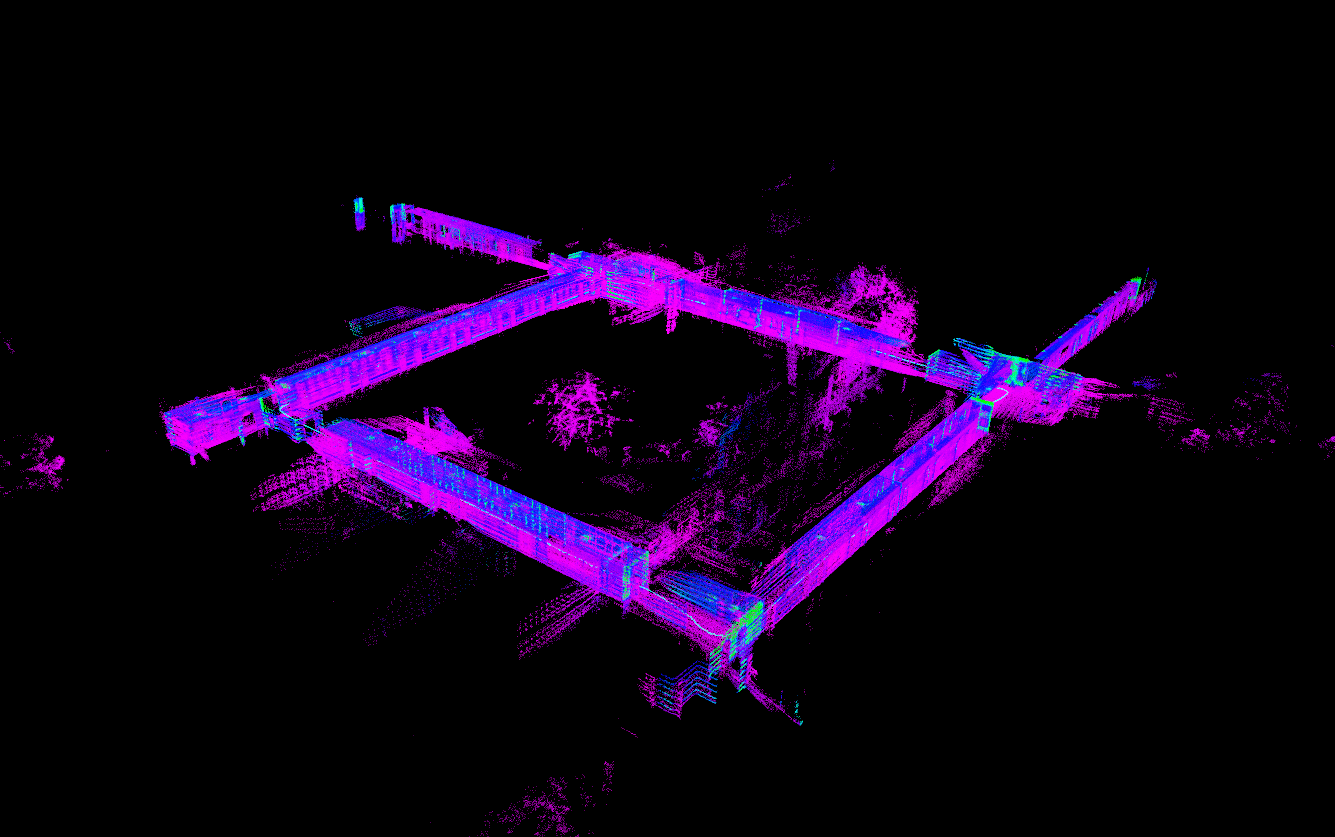}}
\caption{The mapping result in loop1. (a) is the result of VoxelMap, (b) is the result of  VoxelMap++.}
\label{corrider map}
\end{figure}

In the corridor, because of the lack of plane constraints in the forward direction which means that the measurement Jacobian will lack the normal vector like $[1,0,0]$.  
It is inevitable to cause significant cumulative errors in the x-axis of the body frame.
Meanwhile, when turning at an intersection, the LiDAR will scan the next corridor without passing through, which means that the performance of the LIO attitude at this time will largely depend on the forward propagation of the IMU.
This can easily lead to attitude errors, resulting in significant linear cumulative errors.
In VoxelMap++, due to the advantage of plane merging, the entire floor and ceiling will be merged into two large planes, thus constraining the drift on the pitch and avoiding linear cumulative errors which is shown in Fig. \ref{corrider map}.
In fact, this phenomenon is similar to the ground segmentation in LeGO-LOAM, but it is further extended to all coplanar planes. 

As Table. \ref{table corrider} shows, other SLAM algorithms are more prone to cumulative errors in corrider.
Our proposed VoxelMap++ achieves much higher accuracy than others, mainly due to the plane merging can estimate the plane representation more accurately and estimate their covariance in real-time.

% The performance of algorithms is evaluated by end-to-end error, Fig(??)
\begin{table}
\begin{minipage}[htb]{0.5\textwidth}
     \caption{end-to-end error(meters) on indoor corridor sequences}
        \centering
        
        \begin{tabular}{c|ccccc}
        \hline
        \label{table corrider}
        Sequence & loop1 & loop2 & loop3 & loop4 & loop5 \\
        \hline
        LIO-Livox & 7.1192 & 18.5059 & 4.2932 & 10.0270 & 19.6729 \\
        FAST-LIO2 & 2.3812 & 2.8175 & 4.2703 & 13.2844 & 7.1070\\
        Faster-LIO & 2.3488 & 1.1595 & 8.1295 & 1.6455 & 10.1804 \\
        VoxelMap & 9.3457 & 1.1388 & 3.8561 & 5.3355 & 14.2049\\
        VoxelMap++ & \textbf{0.5305} & \textbf{0.5671} & \textbf{0.2543} & \textbf{1.3094} & \textbf{1.4478} \\
        \hline
        Route Length(m) & 202.9 & 219.7 & 315.2 & 317.1 & 405.0    \\
        \hline
      \end{tabular} 
\end{minipage}
\end{table}
\subsection{Consumption of resources}
\label{resources usage}
Another advantage of our proposed VoxelMap++ is less CPU and memory resource usage compared with other state-of-art methods which is shown in Table. \ref{table:resource usage}.
This promotion means our method can be applied to the resource constraint embedding system such as AR equipped with iTOF and low-cost UAV.

\begin{table}[htb]
    \caption{Resources usage (Avg. Comp. Time (ms)/ Mem usage (MB))}
    \centering
    \begin{tabular}{c|cc}
        \hline
     \label{table:resource usage} 
   & small scale (loop1) & large scale (loopE)\\
    \hline
    FAST-LIO2 &  11.5985/201.86  & 35.4858/228.23 \\
    Faster-LIO & 7.3461/135.11 & 15.2910/200.21 \\
    VoxelMap & 5.6830/158.06 & 14.5815/201.22 \\
    VoxelMap++ & \textbf{4.8763/126.45} & \textbf{13.8323/195.91} \\
    \hline
    Route Length(m) & 202.9 &  329.4 \\
     \hline
  \end{tabular} 
\end{table}

From the perspective of memory usage, 
compared with the method based on the point cloud, we only need to maintain a hash table with save the plane information in the value which means we will not maintain millions of point clouds in the map.
Compared with Voxelmap, the plane merging method in \ref{Plane Merging Based on Union-Find} will save the memory usage of voxels as much as possible. 
Ultimately, the map in VoxelMap++ will be represented as several large planes $\mathcal{P}^{k}_i.f$ and dozens of planes $\mathcal{P}^{k}_j$  cannot be merged, rather than hundreds of voxels in VoxelMap.

From the perspective of CPU usage, 
compared with the method based on the point cloud, we use spatial hashing O(1) instead of KNN search O(lg(N)) based on KD-Tree.
The plane parameters can be directly obtained by querying the value in the hash table instead of repeatedly performing plane fitting.
Both these two advantages make the CPU usage of VoxelMap and VoxelMap++ much lower than that of  pointcloud-based methods such as FAST-LIO2.
Compared with VoxelMap, we perform 3DOF plane representation rather than 6DOF which means our proposed VoxelMap++ consumes less computational resources in the propagation of observation noise covariance $\textbf{R}_i$. 
Meanwhile, since all inputs in the plane fitting method \eqref{eq6}\eqref{eq8} are in the form of summation. 
We can incrementally update the value instead of recalculating.
\section{Conclusion and Limitation}
This paper proposes a mergeable voxel mapping method
for online LiDAR(-inertial) odometry. 
Compared with other methods, this method maintains the plane feature with 3DOF representation and corresponding covariance which effectively improves the  calculation speed and saves memory usage.
In order to improve the accuracy of plane fitting, we make full use of the relationship between voxels and then merge the coplanar voxel based on union-find after plane fitting converged.
This paper also shows how to implement the proposed mapping method in an iterated extended Kalman filter-based LiDAR(-inertial) odometry.
The experiment on structured open-source datasets and our own challenging datasets shows that our method can achieve better performance than other state-of-art methods.

However, our method also has some drawbacks. 
For example, the robustness in dynamic scenes, such as closing elevators, will drop significantly.
Thus, we will consider optimizing this method from the perspective of identifying the voxel changing in the future.

% or


\section*{Acknowledgment}
This work was supported by the Science and Technology Project Fund of Sichuan Province, China, under Grant 2018sz0364.
The authors would like to thank HKU-Mars-Lab because of the spirit of open-source. As far as we know, their research has widely influenced academia and industry in China.

\ifCLASSOPTIONcaptionsoff
  \newpage
\fi

\end{document}